\documentclass{article}

\usepackage{geometry}
\usepackage{amsfonts}
\date{}
\author{Khalifeh AlJadda\thanks{Department of Computer Science, 
University of Georgia,
Athens, Georgia. Email: aljadda@uga.edu,jam@cs.uga.edu}\and
Mohammed Korayem\thanks{School of Informatics and Computing,  
Indiana Univeristy, Bloomington, IN.
Email: mkorayem@cs.indiana.edu}\and
Camilo Ortiz\thanks{CareerBuilder, Norcross, GA.
Email: Camilo.Ortiz@careerbuilder.com}\and
Trey Grainger\thanks{CareerBuilder, Norcross, GA.
Email: trey.grainger@careerbuilder.com}\and
John A. Miller\thanks{Department of Computer Science, 
University of Georgia,
Athens, Georgia. Email: jam@cs.uga.edu}\and
William S. York\thanks{Complex Carbohydrate Research Center, 
University of Georgia, Athens, Georgia.
Email: will@ccrc.uga.edu }}

  \usepackage[pdftex]{graphicx}
\usepackage{algpseudocode}
\usepackage{array}
\usepackage[T1]{fontenc}

\usepackage{algorithm}
\begin{document}
%
\title{PGMHD: A Scalable Probabilistic Graphical Model for Massive Hierarchical Data Problems}

\maketitle

\begin{abstract}
In the big data era, scalability has become a crucial requirement for any useful computational model. Probabilistic graphical models are very useful for mining and discovering data insights, but they are not scalable enough to be suitable for big data problems. Bayesian Networks particularly demonstrate this limitation when their data is represented using few random variables while each random variable has a massive set of values. With hierarchical data - data that is arranged in a treelike structure with several levels - one would expect to see hundreds of thousands or millions of values distributed over even just a small number of levels. When modeling this kind of hierarchical data across large data sets, Bayesian networks become infeasible for representing the probability distributions for the following reasons: i) Each level represents a single random variable with hundreds of thousands of values, ii) The number of levels is usually small, so there are also few random variables, and iii) The structure of the network is predefined since the dependency is modeled top-down from each parent to each of its child nodes, so the network would contain a single linear path for the random variables from each parent to each child node. In this paper we present a scalable probabilistic graphical model to overcome these limitations for massive hierarchical data. We believe the proposed model will lead to an easily-scalable, more readable, and expressive implementation for problems that require probabilistic-based solutions for massive amounts of hierarchical data. We successfully applied this model to solve two different challenging probabilistic-based problems on massive hierarchical data sets for different domains, namely, bioinformatics and latent semantic discovery over search logs.
\end{abstract}


%

\section{Introduction}
Probabilistic graphical models (PGM) refer to a family of techniques that merge concepts from graph structures and probability models \cite{1}. They represent the conditional dependencies among sets of random variables \cite{2}.  In the age of big data, PGM’s can be very useful for mining and extracting insights from large-scale and noisy data. The major challenges that PGMs face in this emerging field are the scalability and the restriction that they can only be applied on a propositional domain \cite{15,chickering2004large}. Some extensions have already been proposed to address these challenges, such as hierarchical probabilistic graphical models (HPGM) which aim to extend the PGM to work with non-propositional domains \cite{15,fine1998hierarchical}. The focus of these models is to make Bayesian networks applicable to non-propositional domains, but they do not solve the scalability issues that arise when they are applied to massive data sets.

Massive data sets often exhibit hierarchical properties, where data can be divided into several levels arranged in tree-like structures. Data items in each level depend on or influenced by only the data items in the immediate upper level. For this kind of data the most appropriate PGM to represent the probability distribution would be a Bayesian network (BN), since the dependencies in this kind of data are not bidirectional. A Bayesian network is considered to be feasible when it can provide a concise representation of a large probability distribution where the need cannot be efficiently handled using traditional techniques such as tables and equations~\cite{6}. Such a scenario is not the case with massive hierarchical data, however, since each level only represents one random variable, while the data items in that level are outcomes of that random variable. For example, consider that the hierarchical data are organized as follows: The data items in the top level (root level) represent US cities, while the data items in the second level represent diseases, where each city is connected with the set of diseases that appears in that city. In this case assume we have 19000 cities and 50000 diseases. If we would like to represent this data in a BN, we will consider all of the cities in the root level to be outcomes of one random variable $City$ and all the data items in the second level to be outcomes of another random variable $Disease$. Thus, the BN for this data will be composed of two nodes with single path $City \rightarrow Disease$ while the conditional probability table (CPT) for the $Disease$ will contain 950,000,000 (50000 $\times$ 19000) entries. For this kind of data, we propose a simple probabilistic graphical model (PGMHD) that can represent massive hierarchical data in more efficient way. We successfully apply the PGMHD in two different domains: bioinformatics (for multi-class classification) and search log analytics (for latent semantic discovery).

The main contributions of this paper are as follows:
\begin{itemize}
\item We propose a simple, efficient and scalable probabilistic-based model for massive hierarchical data.
\item We successfully apply this model to the Bioinformatics domain in which we automatically classify and annotate high-throughput mass spectrometry data.
\item We also apply this model for large-scale latent semantic
discovery using 1.6 billion search log entries provided by CareerBuilder.com, using the Hadoop Map/Reduce framework.
\end{itemize}

\section{Background}

Graphical models can be classified into two major categories: (1) directed graphical models, which are often referred to as Bayesian networks, or belief networks, and (2) undirected graphical models which are often referred to as Markov Random Fields, Markov networks, Boltzmann machines, or log-linear models \cite{3}. Probabilistic graphical models (PGMs) consist of both graph structure and parameters. The graph structure represents a set of conditionally independent relations for the probability model, while the parameters consist of the joint probability distributions \cite{1}.
Probabilistic graphical models are often considered to be more convenient than numerical representations for two main reasons \cite{4}:
\begin{enumerate}
\item To encode a joint probability distribution for P($x_1$,...,$x_n$) for $n$ propositional variables with a numerical representation, we need a table with $2^n$ entries.
\item Inadequacy in addressing the notion of independence: to test independence between $x$ and $y$, one needs to test whether the joint distribution of $x$ and $y$ is equal to the product of their marginal probability.
\end{enumerate}

PGMs are used in many domains. For example, Hidden Markov Models (HMM) are considered a crucial component for most of the speech recognition systems~\cite{korayem2007optimizing}. In bioinformatics, probabilistic graphical models are used in RNA sequence analysis~\cite{eddy1994rna}, protein homology detection and sequence alignment~\cite{soding2005protein}, and for genome-wide identification ~\cite{xu2008hmm}. In natural language processing (NLP), HMM and Bayesian models are used for part of speech (POS) tagging~\cite{christodoulopoulos2011bayesian,kupiec1992robust}. The problem with PGMs in general, and Bayesian networks in particular, is that they are not suitable for representing massive data due to the time complexity of learning the structure of the network and the space complexity of storing a network with thousands of random variables. In general, finding a network that maximizes the Bayesian and Minimum Description Length (MDL) scores is an NP-hard problem \cite{5}.


\subsection{Bayesian Networks}
\begin{figure}
\centering
\includegraphics[scale=0.7]{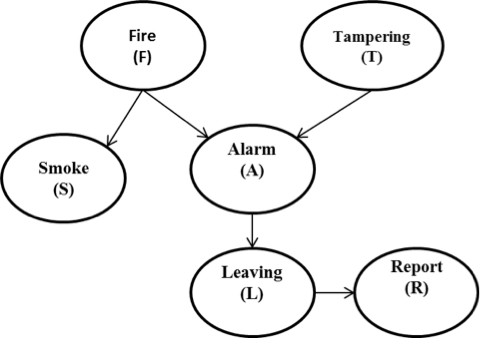}
\caption{Bayesian Network ~\cite{6}}
\label{bn}
\end{figure}

A Bayesian network is a concise representation of a large probability distribution to be handled using traditional techniques such as tables and equations \cite{6}. The graph of a Bayesian network is a directed acyclic graph (DAG)~\cite{2}. A Bayesian network consists of two components: a DAG representing the structure, and a set of conditional probability tables (CPTs) as shown in Figure \ref{bn}. Each node in a Bayesian network must have a CPT which quantifies the relationship between the variable represented by that node and its parents in the network. Completeness and consistency are guaranteed in a Bayesian network since there is only one probability distribution that satisfies the Bayesian network constraints~\cite{6}.
The constraints that guarantee a unique probability distribution are the numerical constraints represented by CPT and the independence constraints represented by the structure itself. The independence constraint is shown in Figure \ref{bn}. Each variable in the structure is independent of any other variables other than its parents, once its parents are known. For example, once the information about A is known, the probability of L will not be affected by any new information about F or T, so we call L independent of F and T once A is known. These independence constraints are known as the Markovian assumptions.

Bayesian networks are widely used for modeling causality in a formal way, for decision-making under uncertainty, and for many other applications \cite{6}.
\subsection{Markov Random Fields (MRFs)}
\begin{figure}
\centering
\includegraphics[scale=0.7]{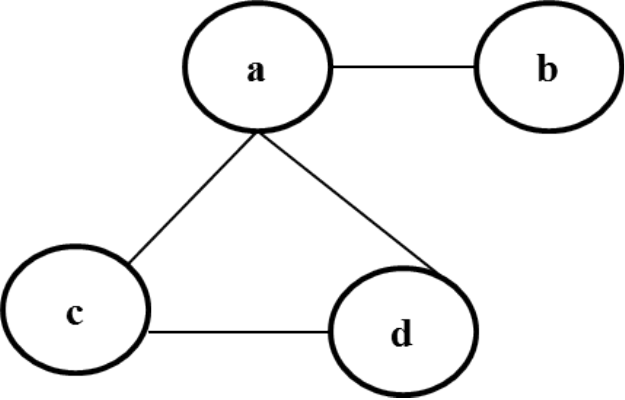}
\begin{center}
\caption{Markov Network}
\label{mn}
\end{center}
\end{figure}

MRFs, which are known also as Markov networks, are the most well-known graphical models in which the graph is undirected. In this graphical model, the random variables are represented as vertices while the edges represent dependency. However, because there is no clear causal influence from one node to the other (i.e. the link represents a direct dependency between two variables, but neither one of them is a cause for the other) the edges are undirected. In an undirected graph any two nodes without a direct link are always conditionally independent variables, whereas any two nodes with a direct link are always dependent \cite{2,4}.
In MRFs the joint probability distribution can be calculated by multiplying a normalization factor by potential functions which assign positive value to a set of fully connected nodes called a clique. A clique is a fully connected subset of nodes that is associated with a non-negative potential function $\phi$. Potential functions are derived from the notion of conditional independence, so any potential function must refer only to the nodes that are directly connected (i.e. form a maximal clique). According to cliques and potential functions, the joint probability in an undirected graph shown in Figure \ref{mn} is calculated using the following equation:
\[p(a,b,c,d)=\frac{1}{Z} \phi _{acd} (a,c,d)\phi _{a,b} (a,b)\]
Where $Z$ is a normalization factor that is calculated by summing or integrating the product of the potential functions:
\[Z=\sum _{a}\sum _{b}\sum _{c}\sum _{d}\phi _{a,c,d} (a,c,d)\phi _{a,b} (a,b)    \]
MRFs are common in many fields like spatial statistics, natural language processing, and communication networks that have little causal structure to guide the construction of a directed graph.
\subsection{Hidden Markov Models}
A Hidden Markov Model (HMM) is a statistical time series model which is used to model dynamic systems whose states are not observable, but whose outputs are. HMMs are widely used in speech recognition, handwriting recognition and text-to-speech synthesis \cite{7}. HMMs rely on three main assumptions. First, the observation at time \textit{t }is generated by a process whose state ${S}_{t}$ is hidden from observation. 

Second, the state of that hidden process satisfies the Markov assumption that once the state of the system at \textit{t} is known, its states and outputs at times after \textit{t} are no longer dependent on states before \textit{t}. In other words, the state at a specific time contains all needed information about the history of the process to predict the future of the process. Upon those assumptions, the joint probability distribution of a sequence of states and observations can be factored as follows \cite{6,7}:
\[P(S_{1:T} ,Y_{1:T} )=P(S_{1} )P(Y_{1} |S_{1} )\prod _{t=2}^{T}P(S_{t} |S_{t-1} )P(Y_{t}  |S_{t} )\]
where ${S}_{t}$ refers to the hidden state, ${Y}_{t}$ refers to the observation at time \textit{t}, and the notation $1:T$ means (1,2,..,T). 

The third assumption is that the hidden state variables are discrete (i.e. S${}_{t}$ can take on $K$ values). So, to define the probability distribution over observation sequences, we need to specify a probability distribution over the initial state \textit{P}(\textit{S}${}_{1}$), the \textit{K}*\textit{K }state transition matrix defining \textit{P}(\textit{S}${}_{t}$\textbar \textit{S}${}_{t-1}$) and the output model defining \textit{P}(\textit{Y}${}_{t}$\textbar \textit{S}${}_{t}$).  HMMs are considered a subclass of Bayesian networks known as dynamic Bayesian networks (DBN), which are Bayesian networks that model systems that evolve over time \cite{8}.

\section{Related Work}

This section describes the most related work to the proposed model from different perspectives. First, we describe the related hierarchical probabilistic models, then we describe the current techniques used to automate the annotation of Mass Spectrometry (MS) data for glycomics, which is one of the scenarios that we use to test the proposed model. We close this section by describing how we applied the proposed model to discover the latent semantic similarity between keywords extracted from search logs for the purposes of building a semantic search system.

\subsection{Probabilistic Graphical Models for Hierarchical Data}
Probabilistic graphical models require propositional domains \cite{15}. To overcome this limitation some extensions were proposed to extend those models to non-propositional domains. A Bayesian hierarchical model has been used for natural scene categorization where it performs well on large sets of complex scenes \cite{fei2005bayesian}.  This model has also been applied for event recognition of human actions and interactions \cite{park2004hierarchical}. Another application of the hierarchical Bayesian network is for identifying changes in gene expression from microarray experiments \cite{broet2002bayesian}

In \cite{15} the authors introduced a hierarchical Bayesian network which extends the expressiveness of a regular Bayesian network by allowing a node to represent an aggregation of simpler types which enables the modeling of complex hierarchical domains. The main idea is to use a small number of hidden variables as a compressed representation for a set of observed variables with the following restrictions:
\begin{enumerate}
\item Any parent of a variable should be in the same or immediate upper layer.
\item At most one parent from the immediate upper layer is allowed for each variable.
\end{enumerate}
So, the idea is mainly to compress the observed data. Although hierarchical Bayesian network models extended the regular Bayesian network to represent non-propositional domains, they have not been able to solve the issue of the scalability of Bayesian networks for massive amounts of hierarchical data.

\subsection{Automated Annotation of Mass Spectrometry Data for Glycomics}
\begin{figure}
\centering
\includegraphics[scale=0.5]{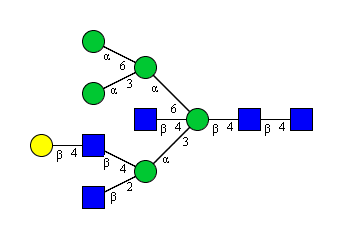}
\caption{Glycan structure in CFG format. The circles and squares represent the monosaccharides which are the building blocks of a glycan while the lines are the linkages between them}
\label{glycan}
\end{figure}

One use case of the proposed model is the automated annotation of Mass Spectrometry (MS) data for glycomics.
Glycans (Figure \ref{glycan}) are the third major class of biological macro-molecules besides nucleic acids and proteins \cite{9}. Glycomics refers to the scientific attempts to characterize and study glycans, as defined in \cite{9} or an integrated systems approach to study structure-function relationships of glycans as defined in \cite{10}.
The importance of this emerging field of study is clear from the accumulated evidence for the roles of glycans in cell growth and metastasis, cell-cell communication, and microbial pathogenesis. Glycans are more diverse in terms of chemical structure and information density than nucleic acids and proteins \cite{10}.
Glycan identification is much more difficult than protein identification, and it is a proven NP-hard problem \cite{11} since, unlike protein structures, glycan structures are trees rather than linear sequences. This leads to a large diversity of glycan structures, which, along with the absence of a standard representation of glycans, has resulted in many incomplete databases, each of which stores glycan structures and glycan-related data in a different format.  For example KEGG \cite{16} uses the KCF format, Glycosciences.de \cite{17} uses the LINUCS format, and CFG \cite{18} uses the IUPAC format.

Although MS has become the major analytical technique for glycans, no general method has been developed for the automated identification of glycan structures using MS and tandem MS data. The relative ease of peptide identification using tandem MS is mainly due to the linear structure of peptides and the availability of reliable peptide sequence databases. In proteomic analyses, a mostly complete series of fragment ions with high abundance is often observed. In such tandem mass spectra, the mass of each amino acid in the sequence corresponds to the mass difference between two high-abundance peaks, allowing the amino acid sequence to be deduced. In glycomics MS data, ion series are disrupted by the branched nature of the molecule, significantly complicating the extraction of sequence information. In addition, groups of isomeric monosaccharides commonly share the same mass, making it impossible to distinguish them by MS alone. Databases for glycans exist but are limited, minimally curated, and suffer badly from pollution from glycan structures that are not produced in nature or are irrelevant to the organism of study.
Several algorithms have been developed in attempts to semi-automate the process of glycan identification by interpreting tandem MS spectra, including CartoonistTwo \cite{13}, GLYCH \cite{20}, GlycoPep ID \cite{21}, GlycoMod \cite{22}, GlycoPeakFinder \cite{23}, GlycoWork-bench \cite{24}, and SimGlycan ~\cite{apte2010bioinformatics} (commercially available from Premier Biosoft). However, each of these programs produces incorrect results when using polluted databases to annotate large MS$^n$ datasets containing hundreds or thousands of spectra. Inspection of the current literature indicates that machine learning and data mining techniques have not been used to resolve this issue, although they have a great potential to be successful in doing so. PGMHD attempts to employ machine learning techniques (mainly probabilistic-based classification) to find a solution for the automated identification of glycans using MS data.

\subsection{Semantic Similarity}

Semantic similarity, which is a metric that is defined over documents or terms in which the distance between them reflects the likeness of their meaning ~\cite{harispe2013semantic}, is well defined in Natural Language Processing (NLP) and Information Retrieval (IR) \cite{mihalcea2006corpus}. Generally there are two major techniques used to compute the semantic similarity: one is computed using a semantic network (Knowledge-based approach) \cite{budanitsky2001semantic}, and the other is based on computing the relatedness of terms within a large corpus of text (corpus-based approach) \cite{mihalcea2006corpus}. The major techniques classified under corpus-based approach are Pointwise Mutual Information (PMI) \cite{bouma2009normalized} and Latent Semantic Analysis (LSA) \cite{dumais2004latent}, though PMI outperform LSA on mining the web for synonyms \cite{turney2001mining}. We applied the proposed PGMHD model to discover related search terms by measuring probabilistic-based semantic similarity between those search terms.

\section{Model Structure}\label{sec:modelStructure}
Consider a (leveled) directed graph $G=(V,A)$ where $V$ and $A\subset V\times V$
denote the sets of nodes and arcs, respectively, such that:
\begin{enumerate}
\item The nodes $V$ are partitioned into $m$ levels $L_{1},\ldots,L_{m}$
and a root node $v_{0}$ such that $V=\cup_{i=0}^{m}L_{i}$, $L_{i}\cup L_{j}=\emptyset$
for $i\not=j$ and $L_{0}=\{v_{0}\}$.
\item The arcs in $A$ only connect one level to the next, i.e., if $a\in A$
then $a\in L_{i-1}\times L_{i}$ for some $i=1,\ldots,m$.
\item An arc $a=(v_{i-1},v_{i})\in L_{i-1}\times L_{i}$ represents the
dependency of $l_{i}$ with its parent $l_{i-1}$, $i=1,\ldots,m$.
Moreover, let $\mbox{pa}:V\to\mathcal{P}(V)$ be a function that given
a node $v$, $\mbox{pa}(v)$ is the set of all its parents, i.e.,
\[
\mbox{pa}(v)=\{w:(w,v)\in A\}.
\]

\item The nodes in each level $L_{i}$ represent all the possible outcomes
of a finite discrete random variable, namely $X_{i}$, $i=1,\ldots,m$.
\end{enumerate}
We now make some remarks about the above assumptions. First, the node
$v_{0}$ in the first level $L_{0}$ can be seen as the root node
and the ones in $L_{m}$ as leaves. Second, an observation $x$ in
our probabilistic model is an outcome of a random variable, namely
$X\in L_{0}\times\cdots\times L_{m}$, defined as 
\[
X=(X_{0}:=v_{0},X_{1},\ldots,X_{m}),
\]
which represents a path from $v_{0}$ to the last level $L_{m}$ such
that $(X_{i-1},X_{i})\in A$ a.s. Third, if $(v_{i-1},v_{i})\not\in A$
for some $v_{i-1}\in L_{i-1}$ and $v_{i}\in L_{i}$, then $P(X_{i-1}=v_{i-1},X_{i}=v_{i})=0$.
In other words, $P(X=x)=0$ whenever $x_{i-1}=v_{i-1}$, $x_{i}=v_{i}$
and $(v_{i-1},v_{i})\not\in A$.

Also, assume that there are $n$ observations of $X$, namely $x^{1},\ldots,x^{n}$,
and let $f:X\times V\to\mathbb{N}$ be a frequency function defined
as 
\[
f(a)=\left|\left\{ x^{j}:(x_{i-1}^{j},x_{i}^{j})=a,\ \mbox{for some }i\in\{1,\ldots,m\}\mbox{ and }j\in\{1,\ldots,n\}\right\} \right|,\qquad\forall a\in V\times V.
\]
Clearly, $f(a)=0$ if $a\not\in A$. These latter observations are
the ones used to train our model.

It should be observed that the proposed model can be seen as a special
case of a Bayesian network by considering a network consisting of
a single directed path with $m$ nodes. However, we believe that a
leveled directed graph that explicitly defines one node per outcome of the random variables (as described above): i) leads to an
easily scalable (and distributable) implementation of the problems
we consider; ii) improves the readability and expressiveness of the
implemented network; and iii) facilitates the training
of the model.

\subsection{Probabilistic-based Classification}

Let $X\in L_{0}\times\cdots\times L_{m}$ be defined as earlier in Section
\ref{sec:modelStructure}. Our model can predict
the outcome at a parent level $i-1$ given an observation%
\footnote{Different from the observations used to train our model.%
} at level $i$ with a classification score. Given $ $an outcome at
level $i$, namely $v_{i}\in L_{i}$, we define the \emph{classification
score} $\mbox{Cl}_{i}(v_{i-1}|v_{i})$ of $v_{i}$ to the parent outcome
$v_{i-1}\in L_{i-1}$ by estimating the conditional probability $P(X_{i-1}=v_{i-1}|X_{i}=v_{i})$
as follows 
\begin{eqnarray*}
\mbox{Cl}_{i}(v_{i-1}|v_{i}) & := & \frac{f(v_{i-1},v_{i})}{T(v_{i})}=\frac{\left(\frac{f(v_{i-1},v_{i})}{T(v_{i-1})}\right)\cdot\left(\frac{T(v_{i-1})}{n}\right)}{\left(\frac{T(v_{i})}{n}\right)}\\
 & \approx & \frac{P(X_{i}=v_{i}|X_{i-1}=v_{i-1})\cdot P(X_{i-1}=v_{i-1})}{P(X_{i}=v_{i})}=P(X_{i-1}=v_{i-1}|X_{i}=v_{i}),
\end{eqnarray*}
where 
\[
T(w)=\sum_{v\in\mbox{pa}(w)}f(v,w),\qquad\forall w\in W.
\]

\subsection{Probabilistic-based Semantic Similarity scoring}

Fix a level $i\in\{1,\ldots.m\}$, and let $X$ and $Y$ be identically
distributed random variables such that $X\in L_{0}\times\cdots\times L_{m}$
is defined earlier in Section \ref{sec:modelStructure}. We define the \emph{probabilistic-based semantic similarity score}
between two outcomes $x_{i},y_{i}\in L_{i}$ by approximating the
conditional joint probability $\mbox{CO}_{i}(x_{i},y_{i}):=P(X_{i}=x_{i},Y_{i}=y_{i}|X_{i-1}\in\mbox{pa}(x_{i}),Y_{i-1}\in\mbox{pa}(y_{i}))$
as
\begin{equation}
\mbox{CO}_{i}(x_{i},y_{i})\approx\prod_{v\in\mbox{pa}(x_{i})}p_{i}(v,x_{i})\cdot\prod_{v\in\mbox{pa}(y_{i})}p_{i}(v,y_{i}),
\end{equation}
where $p_{i}(v,w)=P(X_{i-1}=v,X_{i}=w)$ for every $(v,w)\in L_{i-1}\times L_{i}$.
We can naturally estimate the probabilities $p_{i}(v,w)$ with $\hat{p}(v,w)$
defined as
\[
\hat{p}(v,w):=\frac{f(v,w)}{n}.
\]
Hence, we can obtain the related outcomes of $x_{i}\in L_{i}$ (at
level $i$) by finding all the $w\in L_{i}$ with a large estimated\emph{
}probabilistic-based semantic similarity score $\mbox{CO}_{i}(x_{i},w)$.

\subsection{Progressive Learning}
PGMHD is designed to allow progressive learning. Progressive learning is a learning technique that allows a model to learn gradually over time. Training data does not need to be given at one time to the model, instead the model can learn from any available data and integrate the new knowledge with the represented one. This learning technique is very attractive in the big data age for the following reasons:
\begin{enumerate}
\item Any size of the training data can fit.
\item It can easily learn from new data without the need to re-include the previous training data in the learning.
\item The training session can be distributed instead of doing it in one long-running session.
\item Recursive learning allows the results of the model to be used as new training data, provided they are judged to be accurate by the user. The progressive learning approach for PGMHD is shown in Algorithm 1.
\end{enumerate}  

\begin{figure}
\centering
\includegraphics[scale=0.7]{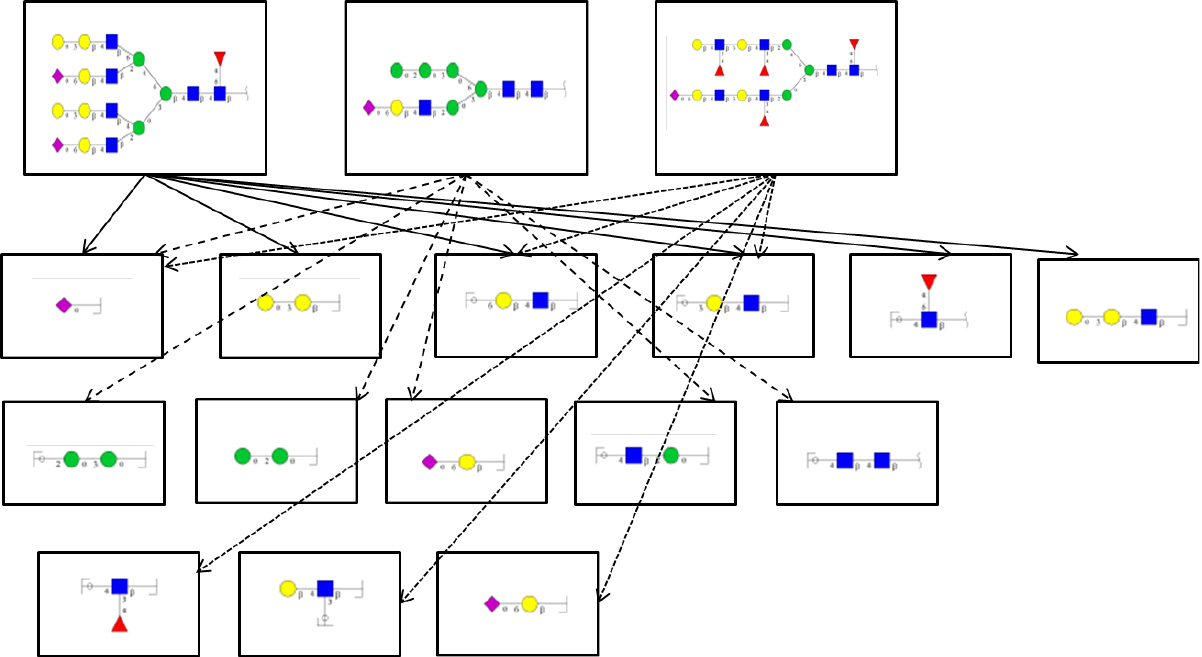}
\caption{PGMHD for tandem MS data. The root nodes are the glycans that annotate the peaks at MS$^1$ level, while the level 2 nodes are the glycan fragments that annotate the peaks at MS$^2$ level and the edges represent dependency between the glycans that generates the fragments.}
\label{pgmhd}
\end{figure}

\begin{algorithm}
\label{alg:learn}
\caption{Progressive Learning for PGMHD}
\includegraphics[scale=1]{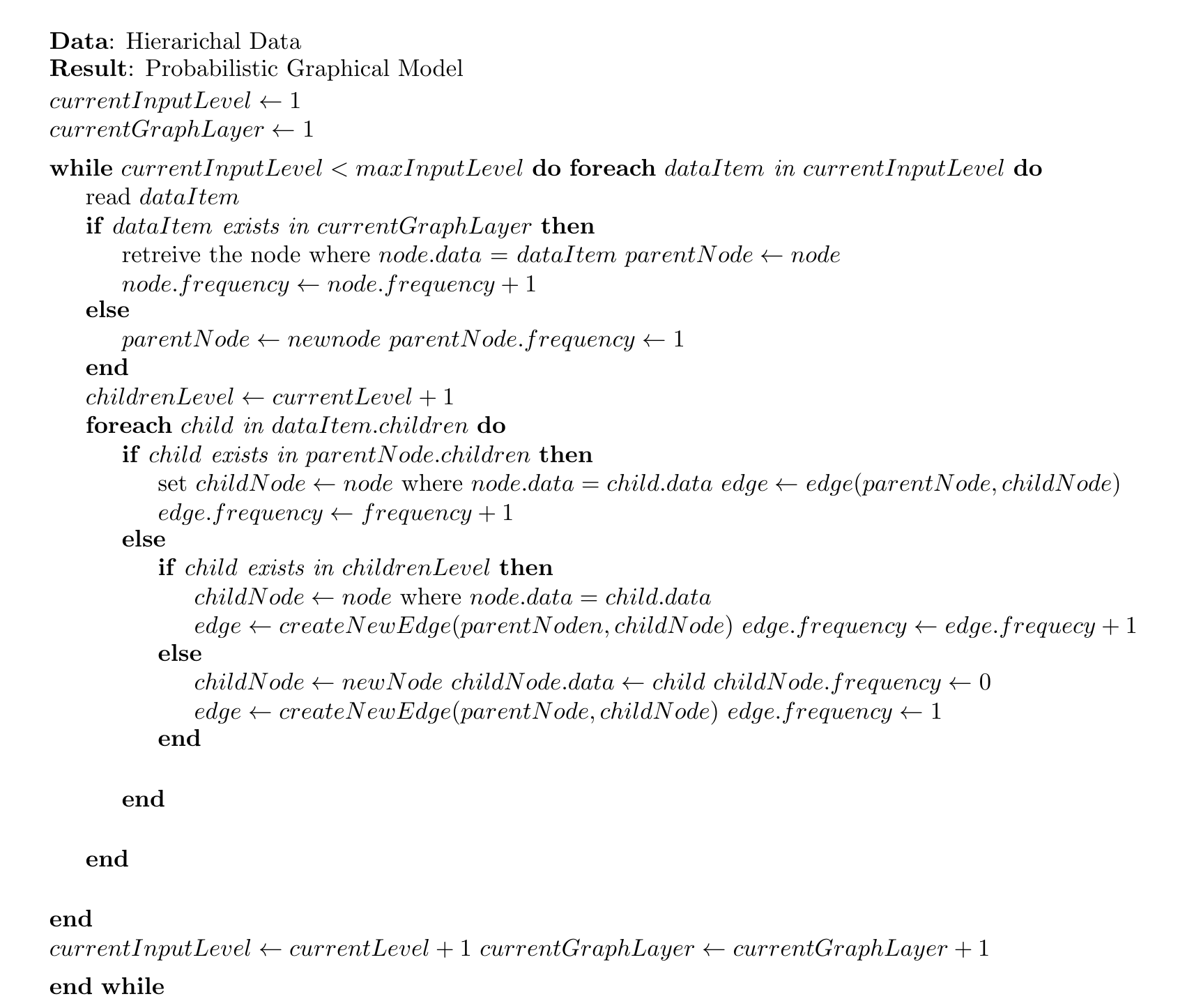}
\end{algorithm}

\section{Experimental Results}

\noindent PGMHD can be used for different purposes once it is built and trained. PGMHD can be used to predict the class from level \textit{l} for the observations of random variables at level \textit{l}+1. For example, in the annotation of the MS data, PGMHD is used to predict the best Glycan at level MS$^1$ to annotate a spectrum by evaluating the annotated peaks at level MS$^2$ with probability scores that represent how well the selected glycan correlates to the manually curated annotations that were used to train the model.

\subsection{PGMHD to automate the MS annotation}
\begin{figure}
\centering
\includegraphics[scale=0.7]{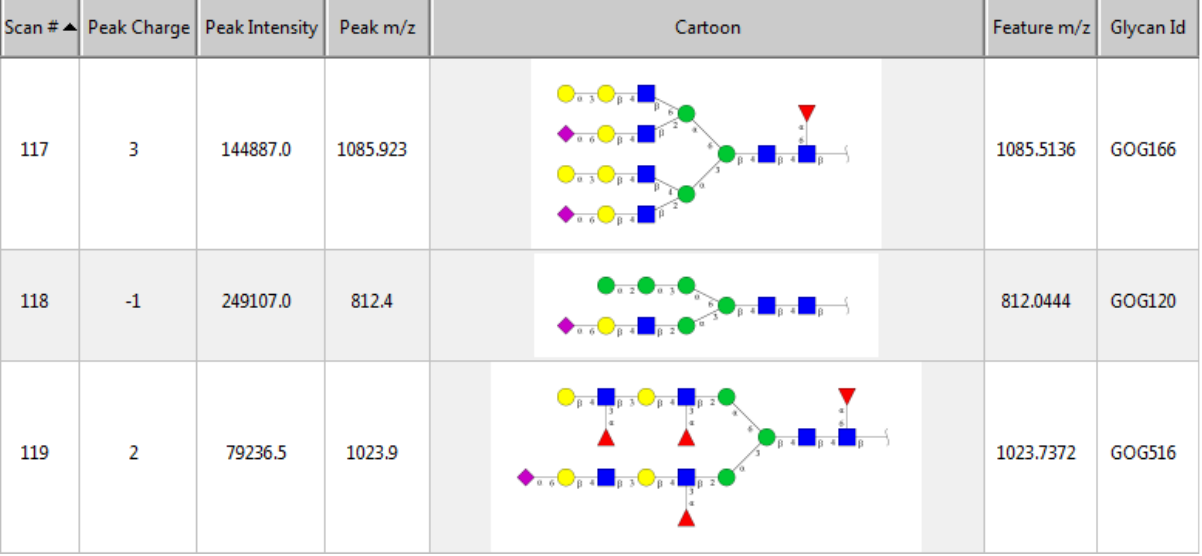}
\caption{MS1 annotation using GELATO. Scan\# is the ID number of the scan in the MS file, peak charge is the charge state of that peak in the MS file, peak intensity represents the abundance of an ion at that peak, peak m/z is the mass over charge of the given peak, cartoon is the annotation of that peak (glycan) in CFG format, feature m/z is the mass over charge for the glycan, and glycanID is the ID of the glycan in the Glycan Ontology(GlycO).}
\label{MS1}
\end{figure}
\begin{figure}
\centering
\includegraphics[scale=0.7]{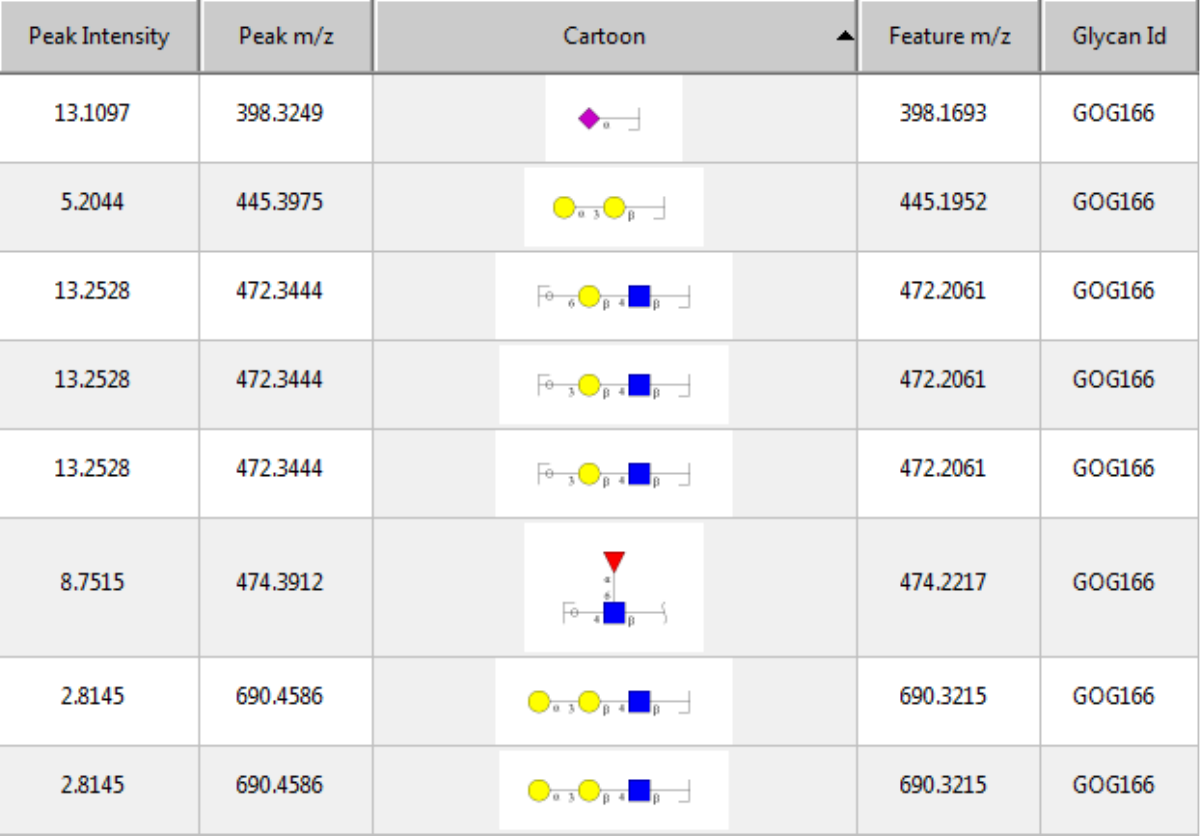}
\caption{Fragments of Glycan GOG166 at the MS$^2$ level. Each ion observed in MS$^1$ is selected and fragmented in MS$^2$ to generate smaller ions, which can be used to identify the glycan structure that most appropriately annotates the MS$^1$ ion. Theoretical fragments of the glycan structure that had been used to annotate the MS$^1$ spectrum are used to annotate the corresponding MS$^2$ spectrum.}

\label{MS2_1}
\end{figure}
\begin{figure}
\centering
\includegraphics[scale=0.7]{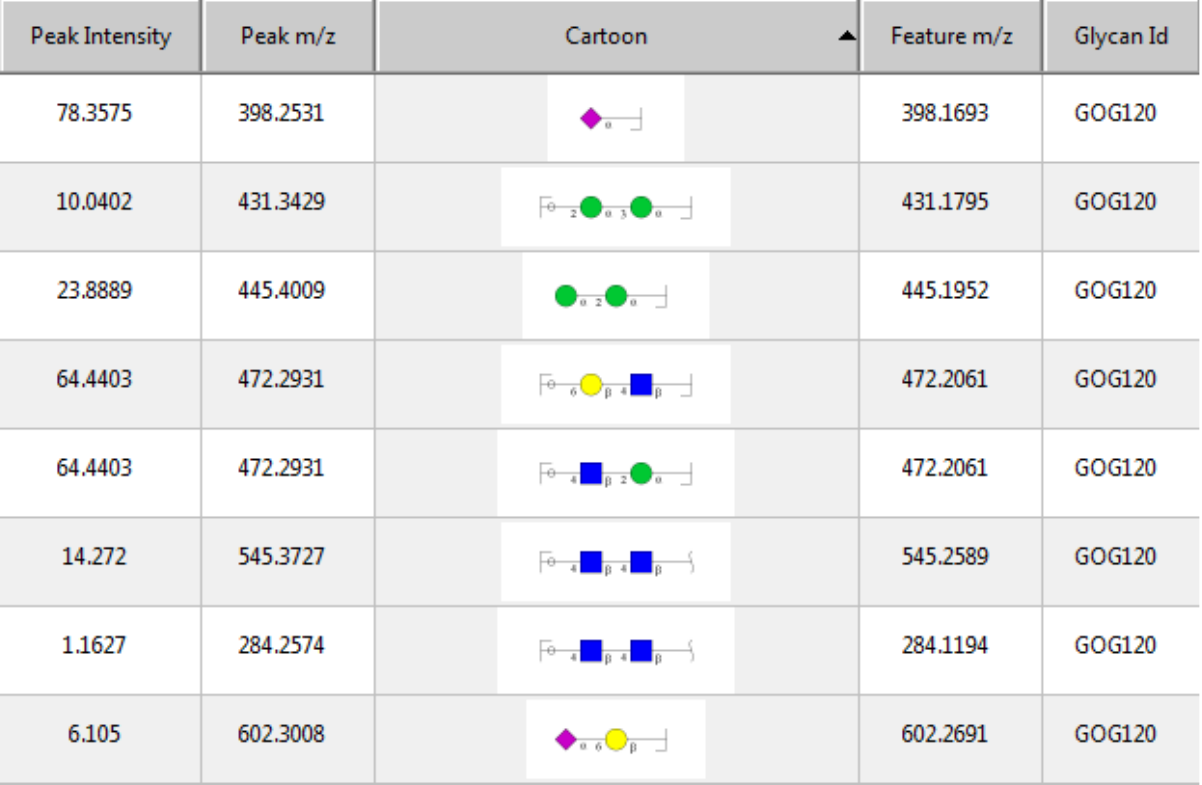}
\caption{Fragments of Glycan GOG120 whose peaks were annotated at the MS$^2$ level. See Figure 5 for annotation scheme.}
\label{MS2_2}
\end{figure}
\begin{figure}
\centering
\includegraphics[scale=0.7]{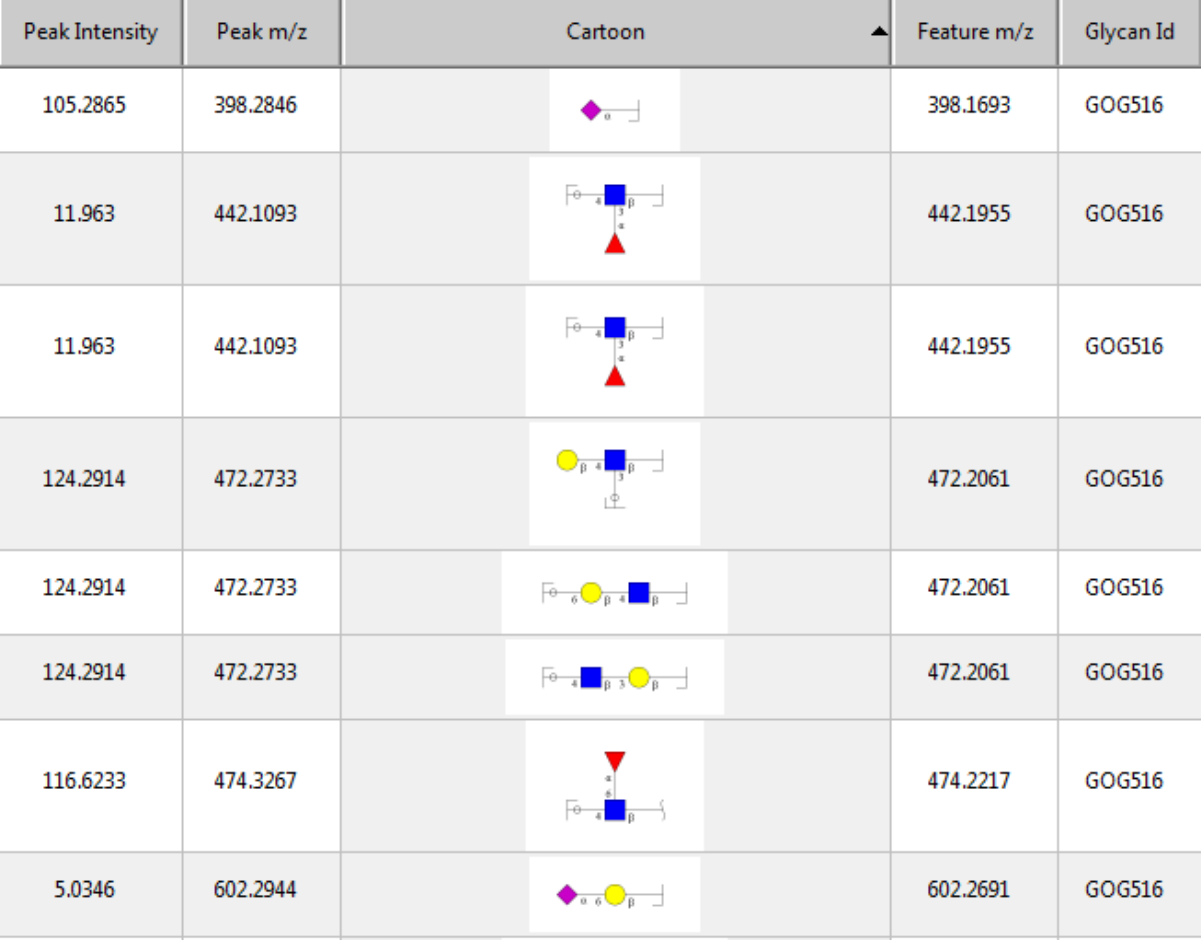}
\caption{Fragments of Glycan GOG516 whose peaks were annotated at the MS$^2$ level. See Figure 5 for annotation scheme.}
\label{MS2_3}
\end{figure}

This model is well suited for representing MS data. We recently implemented the Glycan Elucidation and Annotation Tool (GELATO), which is a semi-automated MS annotation tool for glycomics integrated within our MS data processing framework called GRITS. Figures 4, 5, 6 and 7 show screen shots from GELATO for annotated spectra. Figure \ref{MS1} shows the MS profile level and Figures \ref{MS2_1}, \ref{MS2_2}, and \ref{MS2_3} show the annotated MS$^2$ peaks using fragments of the glycans that were chosen as candidate annotations to the MS profile data (i.e. level 1).

To represent the data shown in these figures using the proposed model, a top-layer node is assigned to each row in the MS profile table, which corresponds to the MS$^1$ data. Then, for each row in the MS$^2$ tables, a unique node is created and connected with its parent node using a directed edge from the parent node (at the MS profile layer) to the child node (at the MS$^2$ layer). Each top-layer node stores a value representing how frequently that parent has been seen in the training data. However, each child node in the MS$^2$ layer has more than one parent. The edge's weight represents the co-occurrence frequency between a child and a parent. The child node stores the total frequency of observing that child regardless of the identity of its parents. The combined frequency data makes it possible to design a progressive learning algorithm that can extract information from massive data sets.
Figure \ref{pgmhd} shows the PGMHD for the given MS data in these figures. As shown in the model, two layers are created: one for the MS$^1$ level and a second one for the MS$^2$ level. The nodes at the MS$^2$ level may have many parents as long as they have the same annotation. The frequency values are not shown because of space constraints.  

We ran our experiments using MS data which is collected from stem cell samples. The size of this data set is 1,746,278 peaks distributed over 1713 MS scans from 10 MS experiments. Figure \ref{learningGraph} shows the learning time using the progressive learning technique. In this test we introduced one new experiment at a time to the model for training, and we recorded the total time required to train the model. These performance results demonstrate how efficiently the progressive learning works with PGMHD.

\begin{table}[!t]
\renewcommand{\arraystretch}{1.3}
\centering
\caption{Precision and Recall for PGMHD in the MS annotation experiment}
\label{pgmhd_prec_recall}
\begin{tabular}{c|c|c}
\hline
\bfseries Size of training set & \bfseries Precision &\bfseries Recall \\
\hline
5 & 0.891 & 0.621\\
\hline
6 & 0.870	& 0.609\\
\hline
7 & 0.865	& 0.619\\
\hline
8 & 0.868	& 0.632\\
\hline
9 & 0.867	& 0.618\\

\end{tabular}
\end{table}

To test the accuracy of PGMHD, we trained the model by randomly selecting one of 10 available experiments, while the other 9 experiments were used to test the trained model by annotating the experiments' peaks using PGMHD. The baseline in our evaluation was the annotations generated by the commercial tool SimGlycan. The results of the accuracy test are shown in Table \ref{pgmhd_prec_recall}. Figure \ref{pgmhd_gelato} shows the average precision and recall for PGMHD compared to the average precision and recall of GELATO using the same dataset of 1,746,278 peaks distributed over 10 MS experiments.
\begin{figure}
\centering
\includegraphics[scale=0.4]{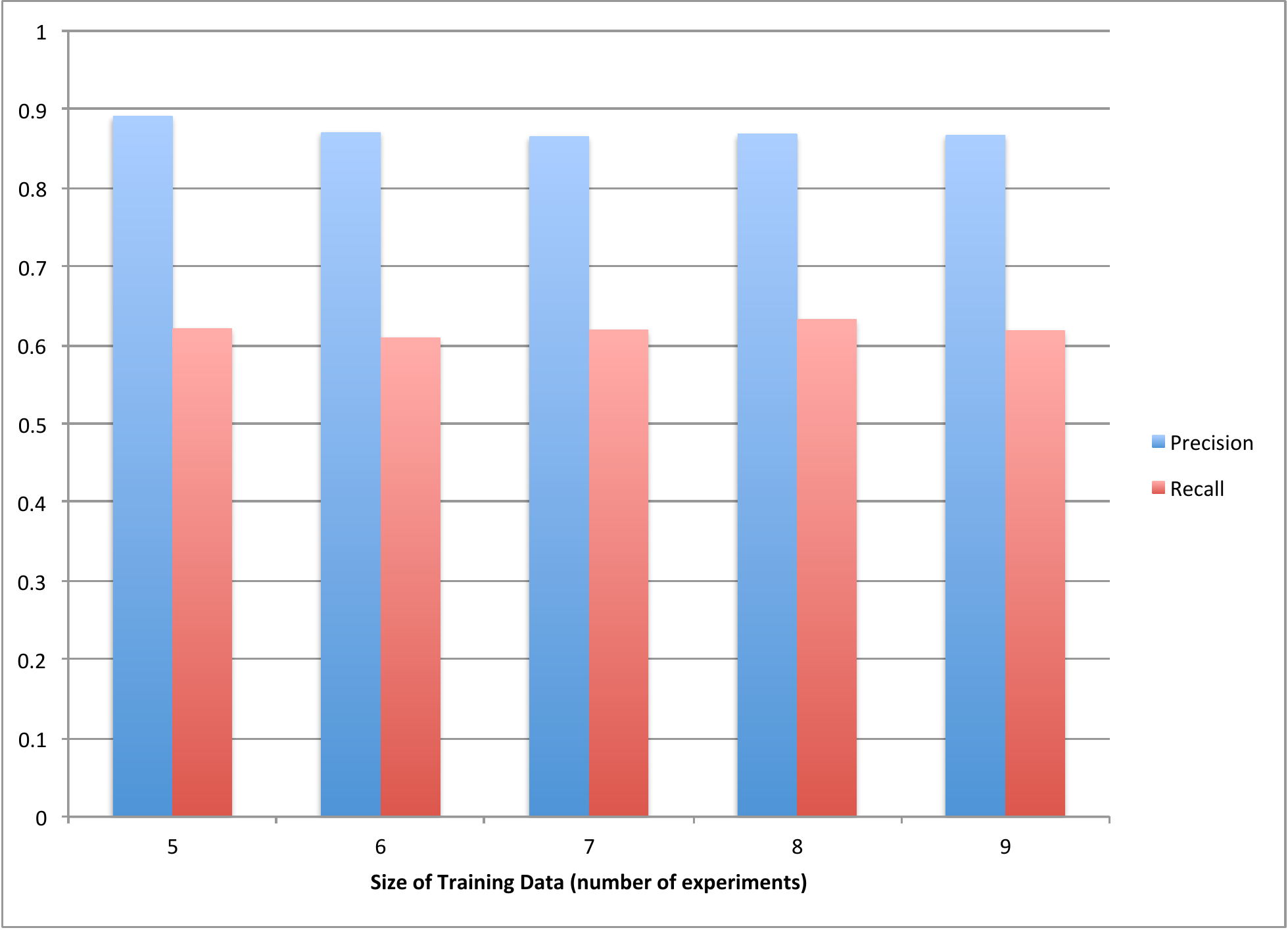}
\caption{Precision and Recall of PGMHD}
\label{prec_glyco}
\end{figure}

\begin{figure}
\centering
\includegraphics[scale=0.7]{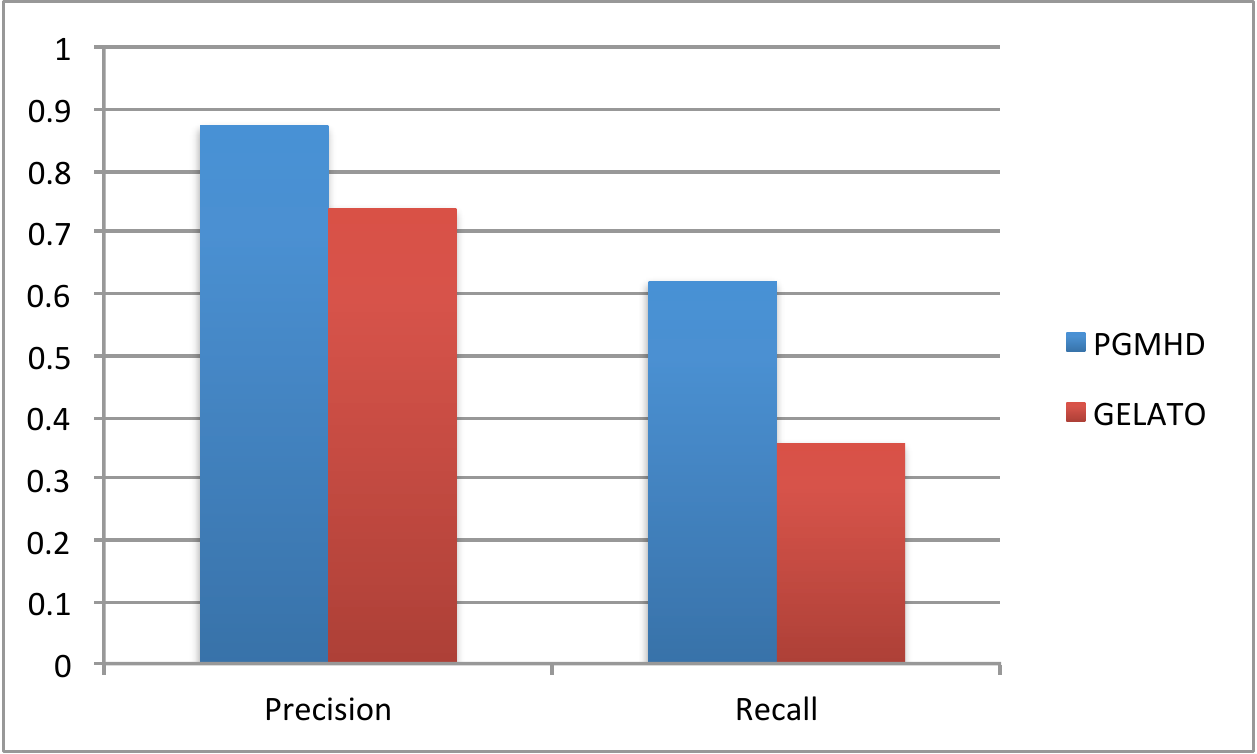}
\caption{Average precision and recall of PGMHD and GELATO}
\label{pgmhd_gelato}
\end{figure}


\begin{figure}
\centering
\includegraphics[scale=0.4]{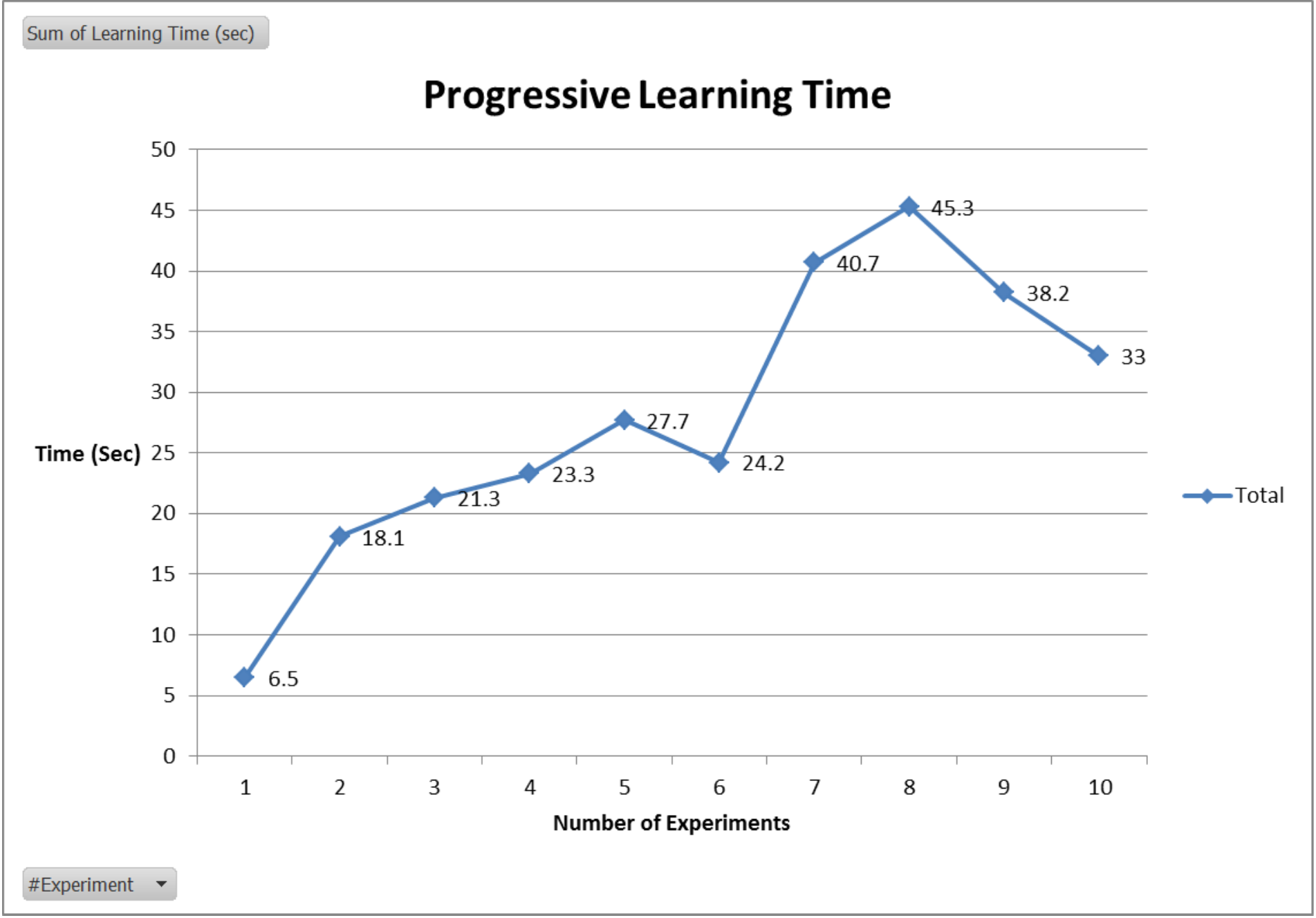}
\caption{Progressive Learning Time Over Different Experiments}
\label{learningGraph}
\end{figure}

\subsection{PGMHD for latent semantic discovery over Hadoop}
We also implemented a version of PGMHD over Hadoop ~\cite{lam2010hadoop} to be used for latent semantic discovery between users' search terms extracted from search logs provided by CareerBuilder.com.
\subsubsection{Problem Description}
CareerBuilder operates the largest job board in the U.S. and has an extensive and growing global presence, with millions of job postings, more than 60 million actively-searchable resumes, over one billion searchable documents, and more than a million searches per hour. The search relevancy and recommendations team wants to discover latent semantic relationships among the search terms entered by their users in order to build a semantic search engine that understands a user's query intent in order to provide more relevant results than a traditional keyword search engine. To tackle this problem, CareerBuilder cannot use a typical synonyms dictionary since most of the keywords used in the employment search domain represent job titles, skills, and companies that would not be found in a traditional English dictionary. Additionally, CareerBuilder's search engine supports over a dozen languages, so they were in search of a model that is language-independent.
\subsubsection{PGMHD over Hadoop}
Given the search logs for all the users and the users' classifications as shown in Table \ref{input_pgmhd}, PGMHD can represent this kind of data by placing the classes of the users as root nodes and placing the search terms for all the users in the second level as children nodes. Then, an edge will be formed linking each search term back to the class of the user who searched for it. The frequency of each search term (how many users search for it) will be stored in the node of that term, while the frequency of a specific search term searched for by users of a specific class (how many users belonging to that class searched for the given term) will be stored in the edge between the class and the term. The frequency of the root node is the summation of the frequencies on the edges that connect that root node with its children (Figure \ref{pgmhdJobs}).
\begin{figure}
\centering
\includegraphics[scale=0.38]{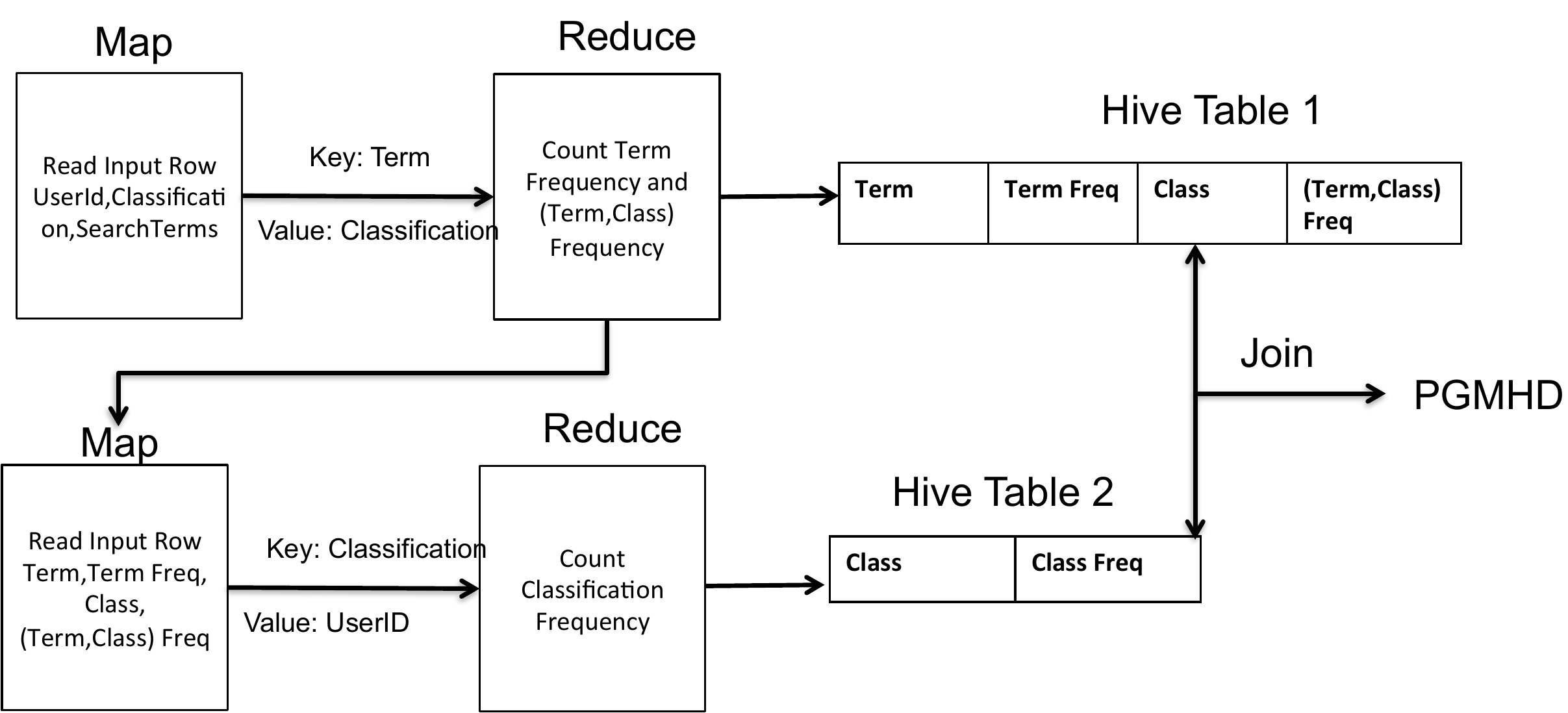}
\caption{PGMHD Over Hadoop}
\label{pgmhdHadoop}
\end{figure}

\begin{table}[!t]
\renewcommand{\arraystretch}{1.3}
\centering
\caption{Input data to PGMHD over hadoop}
\label{input_pgmhd}
\begin{tabular}{c|c|c}
\hline
\bfseries UserID & \bfseries Classification &\bfseries Search Terms \\
\hline
\hline
user1 & Java Developer & Java, Java Developer, C, Software Engineer\\
\hline
user2 & Nurse & RN, Rigistered Nurse, Health Care\\
\hline
user3 & .NET Developer & C\#, ASP, VB, Software Engineer, SE\\
\hline
user4 & Java Developer & Java, JEE, Struts, Software Engineer, SE\\
\hline
user5 & Health Care & Health Care Rep, HealthCare\\
\hline
\end{tabular}
\end{table}
Figure \ref{pgmhdHadoop} shows how PGMHD was implemented over Hadoop using Map/Reduce jobs and Hive tables. After we created PGMHD on Hadoop we calculated the probabilistic-based semantic similarity score between each pair of two terms with shared parents. The size of the data set we analyzed in this experiment is 1.6 billion search records. To decrease the noise in the given data set we applied a pre-filtering technique by removing any search term used by less than 10 distinct users. The final graph representing this data contains 1931 root nodes, 16414 child nodes, and 439435 edges.

\begin{figure}
\centering
\includegraphics[scale=0.4]{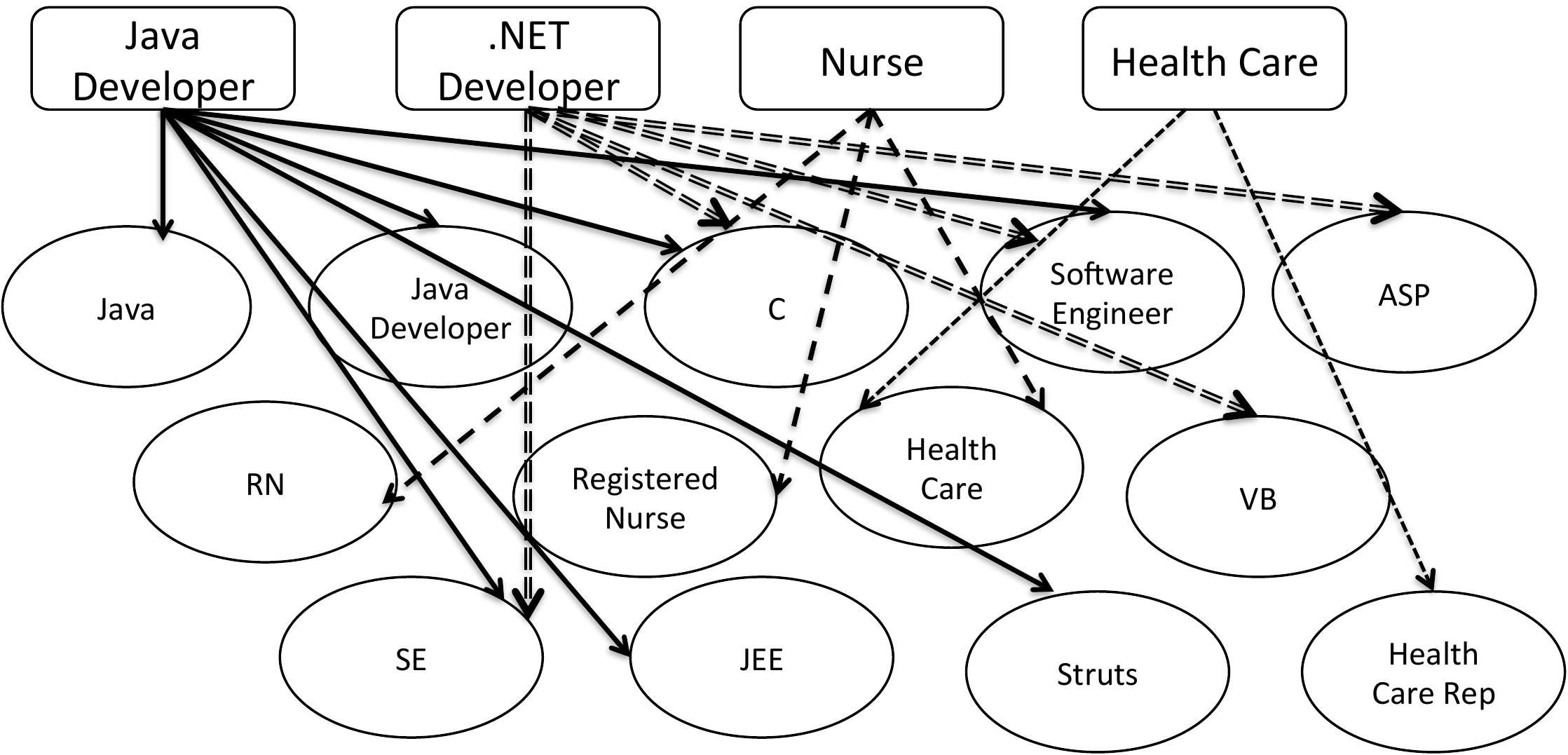}
\caption{PGMHD representing the search log data}
\label{pgmhdJobs}
\end{figure}

\subsubsection{Results of latent semantic discovery using PGMHD}
The experiment performing latent semantic discovery among search terms using PGMHD was run on a Hadoop cluster with 63 data nodes, each having a 2.6 GHZ AMD Opteron Processor with 12 to 32 cores and 32 to 128 GB RAM. Table \ref{pgmhdResults} shows sample results of 10 terms with their top 5 related terms discovered by PGMHD. To evaluate the model's accuracy, we sent the results to data analysts at CareerBuilder who reviewed 1000 random pairs of discovered related search terms and returned the list with their feedback about whether each pair of discovered related terms was ``related" or ``unrelated". We then calculated the accuracy (precision) of the model based upon the ratio of number of related results to total number of results. The results show the accuracy of the discovered semantic relationships among search terms using the PGMHD model to be 0.80.

\begin{table}[!t]
\newcolumntype{L}[1]{>{\raggedright\let\newline\\\arraybackslash\hspace{0pt}}m{#1}}
\newcolumntype{C}[1]{>{\centering\let\newline\\\arraybackslash\hspace{0pt}}m{#1}}
\newcolumntype{R}[1]{>{\raggedleft\let\newline\\\arraybackslash\hspace{0pt}}m{#1}}

\centering
\caption{PGMHD results for latent semantic discovery}
\label{pgmhdResults}
\begin{tabular} {c|L{6cm}}
\hline
Term &  Related Terms\\
\hline
hadoop & big data, hadoop developer, OBIEE, Java, Python\\
\hline
registered nurse & rn registered nurse, rn, registered nurse manager,  nurse, nursing, director of nursing\\
\hline
data mining & machine learning, data scientist,  analytics, business intellegence, statistical analyst\\
\hline
Solr & lucene, hadoop, java\\
\hline
Software Engineer & software developer, programmer, .net developer, web developer, software \\
\hline
big data & nosql, data science, machine learning, hadoop, teradata \\
\hline
Realtor & realtor assistant, real estate, real estate sales, sales,
real estate agent \\
\hline
Data Scientist & machine learning, data analyst, data mining, analytics,
big data\\
\hline
Plumbing & plumber, plumbing apprentice, plumbing maintenance,
plumbing sales, maintenance\\
\hline
Agile & scrum, project manager, agile coach, pmiacp,
scrum master\\

\hline
\end{tabular}
\end{table}

\section{Conclusion}
Probabilistic graphical models are very important in many modern applications such as data mining and data analytics. The major issue with existing probabilistic graphical models is their scalability to handle large data sets, making this a very important area for research given the tremendous modern focus on big data due to the number of data points produced by modern computers systems and sensors. PGMHD is a probabilistic graphical model that attempts to solve the scalability problems in existing models in scenarios where massive hierarchical data is present. PGMHD is designed to fit hierarchical data sets of any size, regardless of the domain to which the data belongs. In this paper we present two experiments from different domains: one being the automated tagging of high-throughput mass spectrometry data in bioinformatics, and the other being latent semantic discovery using search logs from the largest job board in the U.S. The two use cases in which we tested PGMHD show that this model is robust and can scale from a few thousand entries to at least billions of entries, and can also run on a single computer (for smaller data sets), as well as in a parallelized fashion on a large cluster of servers (63 were used in our experiment).


\section*{Acknowledgment}

The authors would like to deeply thank David Crandall from Indiana University for providing very helpful comments and suggestions to improve this paper. We also would like to thank Kiyoko Aoki Kinoshita from Soka University and Khaled Rasheed from University of Georgia for the valuable discussions and suggestions to improve this model. Deep thanks to Melody Porterfield and Rene Ranzinger from Complex Carbohydrate Research Center (CCRC) at the University of Georgia for providing the MS data and the valuable time and information they shared with us to understand the annotation process of the MS data.



%




\bibliographystyle{abbrv}
\bibliography{PGMHD-Arxiv}

\end{document}